\PassOptionsToPackage{vietnamese,english}{babel}

\documentclass{article}

\usepackage[english]{babel}

\usepackage[letterpaper,top=2cm,bottom=2cm,left=3cm,right=3cm,marginparwidth=1.75cm]{geometry}
\usepackage[utf8]{inputenc}
\usepackage{biblatex} 
\usepackage{amsmath}
\usepackage{graphicx}
\usepackage[colorlinks=true, allcolors=blue]{hyperref}
\usepackage{tcolorbox}
\usepackage{enumitem}
\usepackage{authblk}
\usepackage{csquotes}
\usepackage{makecell}
\usepackage{tabularx}

\title{ ViLLM-Eval: A Comprehensive Evaluation Suite for Vietnamese Large Language Models}

\author[1]{Trong-Hieu Nguyen}
\author[1]{Anh-Cuong Le}
\author[2,3]{Viet-Cuong Nguyen}
\affil[1]{Natural Language Processing and Knowledge Discovery Laboratory, Faculty of Information Technology, Ton Duc Thang University, Ho Chi Minh City, Vietnam}
\affil[2]{Intelligent Integration Co., Ltd ($\text{INT}^{2}$), Ha Noi, Vietnam}
\affil[3]{HPC SYSTEMS Inc., Tokyo, Japan}

\begin{document}
\maketitle

\begin{abstract}
The rapid advancement of large language models (LLMs) necessitates the development of new benchmarks to accurately assess their capabilities. To address this need for Vietnamese, this work aims to introduce ViLLM-Eval, the comprehensive evaluation suite designed to measure the advanced knowledge and reasoning abilities of foundation models within a Vietnamese context. ViLLM-Eval consists of multiple-choice questions and predict next word tasks spanning various difficulty levels and diverse disciplines, ranging from humanities to science and engineering. A thorough evaluation of the most advanced LLMs on ViLLM-Eval revealed that even the best performing models have significant room for improvement in understanding and responding to Vietnamese language tasks. ViLLM-Eval is believed to be instrumental in identifying key strengths and weaknesses of foundation models, ultimately promoting their development and enhancing their performance for Vietnamese users\footnote{The ViLLM-Eval data and evaluation code are available at \href{ https://huggingface.co/datasets/vlsp-2023-vllm/ViLLM-Eval}{ https://huggingface.co/datasets/vlsp-2023-vllm/ViLLM-Eval}}. This paper provides a thorough overview of ViLLM-Eval as part of the Vietnamese Large Language Model shared task, held within the 10th International Workshop on Vietnamese Language and Speech Processing (VLSP 2023)\footnote{\href{https://vlsp.org.vn/vlsp2023}{https://vlsp.org.vn/vlsp2023}}.
\end{abstract}

\section{Introduction}

Evaluation benchmarks play a pivotal role in the development of artificial intelligence (AI) systems. Traditionally, natural language processing (NLP) benchmarks have primarily focused on assessing specific and relatively straightforward abilities. However, the advent of large language models (LLMs), also known as foundation models, has brought about a paradigm shift. These powerful models have demonstrated a wide array of novel capabilities, prompting a redirection in the evaluation focus towards more general and intricate skills, such as comprehensive world knowledge and complex reasoning abilities.

To align with the remarkable advancements in LLMs, new benchmarks have emerged to probe the diverse and multifaceted capabilities of these models. For instance, MMLU \cite{hendrycks2021measuring}, HellaSwag \cite{zellers2019hellaswag}, ARC \cite{allenai:arc}, and TruthfulQA \cite{lin2022truthfulqa} are benchmark datasets that have garnered widespread recognition among researchers and are frequently employed on leaderboards to evaluate the performance of language models. However, these benchmarks are primarily tailored to the English language, resulting in a limited understanding of LLMs' capabilities in other languages, including Vietnamese.

Despite the recent surge in powerful Vietnamese LLMs, such as Vistral-7B-Chat \cite{chien2023vistral}, PhoGPT-4B-Chat \cite{PhoGPT}, and VinaLLaMA-7B-Chat \cite{nguyen2023vinallama}, benchmarking these models on datasets translated from English to Vietnamese, even with perfect translations, cannot adequately assess the true quality of these language models concerning their knowledge about core interests of Vietnamese users. This includes aspects of Vietnamese culture, history, and law, as well as other capabilities unique to the Vietnamese society. Conversely, English benchmarks tend to exhibit geographical biases towards the domestic knowledge of the regions that produce them, which may not fully capture the nuances and complexities of other linguistic and cultural contexts.

In this paper, we introduce ViLLM-Eval, a comprehensive evaluation suite meticulously designed to thoroughly assess the advanced knowledge and reasoning abilities of foundation models within the Vietnamese context, as part of the Vietnamese Large Language Models challenge at VLSP 2023. ViLLM-Eval comprises multiple-choice questions and predict-next-word tasks spanning various difficulty levels and diverse disciplines, ranging from humanities to science and engineering. The questions have been carefully curated to reflect the knowledge and reasoning skills relevant to Vietnamese users, including aspects of Vietnamese culture, history, and current affairs.

A comprehensive evaluation of the most advanced Vietnamese LLMs was conducted on ViLLM-Eval. The results revealed that even the best-performing models have significant room for improvement in understanding and responding to Vietnamese language tasks. This finding underscores the pressing need for further development and fine-tuning of LLMs specifically tailored to the Vietnamese language.

By introducing ViLLM-Eval, this work aims to establish a robust and contextually relevant benchmark for evaluating the capabilities of Vietnamese LLMs. This endeavor not only contributes to the advancement of natural language processing in the Vietnamese linguistic landscape but also paves the way for more inclusive and culturally nuanced AI systems that can better serve diverse communities worldwide. The development of language models that truly understand and engage with the intricacies of local contexts is crucial for fostering trust, acceptance, and the equitable distribution of the benefits of AI technologies across societies.

\section{Related Work}

For English benchmarks, traditional English benchmarks mainly focus on assessing certain abilities of models on a single task or a single type of tasks, such as natural language understanding \cite{wang2019glue}, reading comprehension , machine translation \cite{ws-2014-statistical}, and summarization \cite{narayan-etal-2018-dont}. As a representative example, the GLUE benchmark \cite{wang2019glue} combines a collection of NLU tasks, and has witnessed superhuman model performance due to the burst of pretraining models such as BERT \cite{devlin2019bert} and GPT \cite{Radford2019LanguageMA}. In order to assess the capabilities of LLMs more comprehensively, recent benchmarks have cast light on the broader knowledge and advanced abilities. The MMLU benchmark \cite{hendrycks2021measuring} provides multi-domain and multi-task evaluation collected from real-world examinations and books. LLMs’ performance on MMLU fluctuates around random-chance accuracy until they reach the scale of GPT-3 \cite{brown2020language}. The BIG-bench benchmark \cite{suzgun2022challenging} consists of 204 diverse tasks, some of which are considered to be beyond the capabilities of current LLMs.

For Vietnamese benchmarks, while English benchmarks have seen significant development, the assessment of language models in the Vietnamese context is still in its infancy. The focus of Vietnamese benchmarks has largely been on specialized areas, such as the UIT-ViNewsQA \cite{vannguyen2021new}, which is dedicated to understanding health-related information. The same team has also introduced the UIT-ViQuAD \cite{vannguyen2020vietnamese}, a detailed dataset aimed at testing machine reading comprehension specifically in Vietnamese, thus providing a means to gauge how well models comprehend texts in Vietnamese. Additionally, the UIT-ViWikiQA \cite{do2021sentence} is centered around sentence extraction for machine reading comprehension, showcasing a shift towards tackling more complex comprehension challenges within the Vietnamese language. In a significant stride towards a more holistic evaluation of language models in Vietnamese, ZaloAI-Jaist introduced the Vietnamese Multitask Language Understanding Benchmark Suite in 2023 \cite{Charles2013}. Additionally, this benchmark suite enriches its evaluation criteria by incorporating a diverse range of multiple-choice questions, spanning various difficulty levels, including university-level questions across numerous advanced disciplines. This inclusion aims to rigorously test the models' understanding and reasoning in specialized and higher education contexts, significantly broadening the scope of Vietnamese language model assessment.

\newpage
\section{Methodology}

\subsection{Design Principle}

The ViLLM-Eval suite is specifically designed to assess and quantify the knowledge and contextual reasoning capabilities of Large Language Models (LLMs), with a focus on four distinct evaluation dataset, each governed by unique criteria:
\begin{itemize}
    \item LAMBADA\_vi dataset: This dataset is similar to the LAMBADA dataset \cite{paperno2016lambada} using complexity metrics to evaluate LLM's contextual reasoning ability. It challenges the model to predict the final word of a sentence within a paragraph consisting of 4-6 sentences. This test underscores the model's ability to leverage contextual clues distributed across multiple sentences, providing insight into its depth of linguistic comprehension.
    \item Exam Vietnamese dataset: The diverse set of multiple choice questions spans grades 6 to 12, this dataset includes subjects such as Math, Physics, Chemistry, Biology, History, Geography and Literature, obviously excluding includes English. The accuracy metric is utilized here to gauge specialized knowledge across these various domains, offering a measure of the model's academic proficiency. This dataset uses the same format, similar to MMLU \cite{hendrycks2021measuring}.
    \item General Knowledge dataset: This segment amalgamates multiple-choice questions derived from popular Vietnamese television shows like "Who Wants to Be a Millionaire" and "Road to the Olympia Peak." These questions cover a broad spectrum of general knowledge topics related to Vietnam and the world at large, testing the model's breadth of worldly awareness. This dataset uses the same format, similar to MMLU \cite{hendrycks2021measuring}.
    \item Comprehension QA dataset: The final dataset of the suite consists of multiple-choice comprehension questions based on lengthy text passages. This dataset simulates a more in-depth reading comprehension scenario, requiring the model to parse and understand extensive narratives before selecting the correct answer from four options. This dataset uses the same format, similar to MMLU \cite{hendrycks2021measuring}.
\end{itemize}
The selection of perplexity and accuracy as the primary metrics for these evaluations is strategic. Despite their simplicity, these metrics offer a robust and straightforward means of assessing LLMs. Perplexity provides a quantitative measure of the model's ability to predict and understand language within a given context, while accuracy offers a direct evaluation of the model's knowledge and reasoning capabilities across a wide range of subjects and scenarios. Together, these metrics furnish a comprehensive picture of an LLM's linguistic and cognitive faculties, enabling targeted improvements and enhancements.

\begin{table}[h]
    \centering
    \begin{tabularx}{\textwidth}{|c|c|X|} \hline 
         Name dataset& Num samples &Description\\ \hline 
         LAMBADA\_vi& 10,246 & The LAMBADA dataset evaluates LLMs' contextual reasoning ability by challenging them to predict the final word of a sentence within a paragraph consisting of 4-6 sentences.\\ \hline 
         Exam& 19,150 & The Exam Vietnamese dataset includes multiple-choice questions spanning various subjects from grades 6 to 12, providing insight into the model's academic proficiency across diverse domains.\\ \hline 
         Gen. Knowl.& 2,000 & The General Knowledge dataset comprises multiple-choice questions derived from popular Vietnamese television shows, testing the model's breadth of worldly awareness.\\ \hline 
         Comp. QA&900 & The Comprehension QA dataset simulates an in-depth reading comprehension scenario, requiring the model to parse and understand extensive narratives before selecting the correct answer.\\ \hline
    \end{tabularx}
    \caption{This table presents the number of samples contained in various evaluation datasets used to assess the performance of language models. The datasets listed include LAMBADA, Exam, General Knowledge (Gen. Knowl.), and Comprehensive QA (Comp. QA). Having information about the sample sizes of these datasets is important for understanding the scope and scale of the evaluation process.}
    \label{tab:eval_dataset_samples}
\end{table}

\subsection{\textbf{LAMBADA\_vi}}

\subsubsection{Data sources}
The data for the LAMBADA\_vi challenge was sourced exclusively from the Vietnamese Wikipedia. This choice ensures a rich diversity of topics and linguistic structures, mirroring real-world usage of the Vietnamese language. Wikipedia provides a comprehensive and ever-expanding repository of knowledge, making it an ideal source for challenging language models with complex contextual understanding tasks.

\subsubsection{Data Processing}

\begin{figure}
    \centering
    \includegraphics[width=1\linewidth]{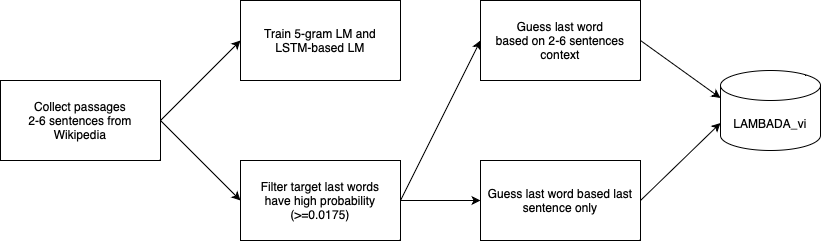}
    \caption{Process of creating LAMBADA\_vi dataset}
    \label{fig:lambada-process}
\end{figure}

The processing of the data involved several meticulous steps to ensure the quality and relevance of the text similar to the LAMBADA dataset \cite{paperno2016lambada}. Initially, the raw text was segmented into paragraphs and sentences. Only paragraphs containing 2 to 6 sentences were retained for further processing. Paragraphs with fewer than 3 sentences were discarded due to their limited contextual scope, whereas paragraphs exceeding 6 sentences underwent random sentence removal from either the beginning or the end to conform to the 6-sentence limit.

Subsequently, the selected text segments were subjected to a filtering process using two models: an Ngram Language Model (LM) utilizing kenlm \cite{heafield-2011-kenlm} and an LSTM-based LM \cite{10.1162/neco.1997.9.8.1735}. These models were employed to eliminate paragraphs where the predictability of the final word exceeded a probability threshold of 0.0175. Both filtering models were either statistically compiled or trained on a corpus combining original Wikipedia data with additional texts from CommonCrawl, ensuring a broad linguistic foundation.

The final step in data processing involved a manual review by a team of evaluators. This review aimed to refine the selection further by ensuring that the context provided by the preceding sentences within a paragraph was necessary to predict the final word accurately. Paragraphs where the last word could be easily predicted without the contextual buildup or, conversely, where the final word remained unpredictable despite adequate contextual information, were excluded from the dataset. This rigorous process ensured that the remaining text segments met the specific challenge criteria, emphasizing the importance of contextual understanding in language processing.

\begin{tcolorbox}[colback=white,colframe=black,boxrule=0.25pt]
\selectlanguage{vietnamese}
\textbf{Example:} Thế kỷ thứ 6, Sur là trung tâm buôn bán với các nước Đông Phi. Trong thế kỷ thứ 16, thành phố này dưới sự cai trị của Bồ Đào Nha, nhưng được imam (nhà lãnh đạo Hồi giáo) người Oman là Nasir ibn Murshid giải phóng. Sau đó thành phố trải qua một cuộc phục hưng kinh tế và là trung tâm buôn bán với Ấn Độ và các nước .......... -> \textbf{Đông Phi}

\textbf{English translate:} \textit{In the 6th century, Sur was a trading hub with East African countries. In the 16th century, the city was under the rule of Portugal, but was liberated by Nasir ibn Murshid, an Omani Imam. Subsequently, the city experienced an economic revival and became a trading center with India and ......... countries. -> \textbf{East African}}
\end{tcolorbox}
\begin{tcolorbox}[colback=white,colframe=black,boxrule=0.25pt]
\selectlanguage{vietnamese}
\textbf{Example:} Hủ tiếu gõ ít khi bán buổi sáng mà thường từ khoảng 14-15 giờ chiều cho đến tận khuya. Nghề bán hủ tiếu gõ tuy không quá nặng nhọc nhưng cũng không nhẹ nhàng, người bán phải đi chợ từ sớm, chuẩn bị mọi thứ cho việc bán buổi chiều, rồi thức đêm bởi đây là món ăn khuya. Để có một gánh hủ tiếu bán đêm thì ngay sáng sớm họ phải len lỏi qua các chợ đầu mối để mua nguyên vật liệu từ thịt, giá, hẹ, tương, ớt, chanh... tất bật từ sáng đến trưa để chuẩn bị đầy đủ mọi thứ. Bán hủ tiếu gõ không cần khéo tay, cũng không cần nhiều .......... -> \textbf{vốn}

\textbf{English translate:} \textit{Hủ tiếu gõ is rarely sold in the morning but usually from around 2-3 p.m. until late at night. Selling hủ tiếu gõ is not overly laborious but neither is it light; vendors have to go to the market early to prepare everything for the afternoon sales, and then stay up late as this is a late-night dish. To have a cart of hủ tiếu gõ for sale at night, they must sneak through wholesale markets early in the morning to buy ingredients such as meat, noodles, onions, garlic, chili, lime... bustling from morning until noon to prepare everything. Selling hủ tiếu gõ does not require skilled hands, nor does it require much .......... -> \textbf{capital}}
\end{tcolorbox}

\subsection{\textbf{Exam Vietnamese Dataset}}

\subsubsection{Data sources}
This work aims to collect primary data for ViLLM-Eval from freely available mock exams on the internet. These include past exam papers from previous years that are publicly shared for students by schools or educational institutions, as well as exam questions compiled from question banks of reputable educational websites in Vietnam. The Exam Vietnamese Dataset covers three difficulty levels: middle school, high school, and general. It includes standard subjects for middle school level and high school level in Vietnam, except for the English subject. Subjects include Mathematics, Physics, Chemistry, Biology, History, Geography, Literature, and more. To ensure accuracy and reliability of the data, it is important to collect from credible sources such as the website of the Ministry of Education and Training of Vietnam and popular and trusted educational websites in Vietnam. Additionally, a small portion of questions may be collected from copyrighted sources, provided that permission and confirmation are obtained from the respective copyright holders. Utilizing free and credible data sources will help ensure the quality and accuracy of the data in ViLLM-Eval, while adhering to copyright and intellectual property regulations.

\subsubsection{Data processing}
This work aims to collect data in various formats, primarily from web pages, and a minor fraction from PDF and Microsoft Word documents. All questions are subsequently parsed – automatically when possible, and otherwise manually by the authors – into a structured format, similar to MMLU \cite{hendrycks2021measuring}. For subjects in the STEM category with complex mathematical notations, they are manually converted into standard LaTeX formats. All questions in ViLLM-Eval are processed to include exactly four answer choices. Questions containing images are eliminated, as this evaluation focuses on LLMs, and those with answer choices such as "All of the above are correct," "All of the above are incorrect," "Options A and B are correct," "Options A and C are incorrect," etc., are removed as these answer choices are not compatible with the evaluation method. All questions go through the standard data preprocessing pipeline, such as deduplication and cleaning. Following this, the questions undergo several rounds of human validation by the authors, and all LaTeX notations are ensured to be compliant without syntax errors.

\begin{tcolorbox}[colback=white,colframe=black,boxrule=0.25pt]
\selectlanguage{vietnamese}
\textbf{Example:} Để tạo ra các giống vật nuôi có tốc độ sinh trưởng và phát triển nhanh, năng suất cao, thích nghi với các điều kiện địa phương, người ta áp dụng các phương pháp? (Môn Sinh)
\textit{To create livestock breeds with fast growth and development rates, high productivity, and adaptation to local conditions, which methods are applied?}

\begin{enumerate}[label=(\alph*)]
    \item Xây dựng và cải tạo chuồng trại trong chăn nuôi \textit{Constructing and improving animal husbandry facilities}
    \item Đảm bảo vệ sinh, phòng trừ dịch bệnh cho vật nuôi \textit{Ensuring hygiene and disease prevention for livestock}
    \textbf{\item Chọn lọc nhân tạo, lai giống, công nghệ tế bào \textit{Artificial selection, crossbreeding, and cell technology}}
    \item Cải tạo chế độ dinh dưỡng, đáp ứng đầy đủ nhu cầu về thức ăn \textit{Improving nutrition regimes to meet the full dietary needs}
\end{enumerate}
\end{tcolorbox}

\begin{tcolorbox}[colback=white,colframe=black,boxrule=0.25pt]
\selectlanguage{vietnamese}
\textbf{Example:} Cơ năng của con lắc lò xo dao động điều hòa không phụ thuộc vào? (Môn Vật lí)
\textit{The oscillatory motion of a harmonic oscillator does not depend on:}

\begin{enumerate}[label=(\alph*)]
    \item Độ cứng của lò xo \textit{The stiffness of the spring}
    \item Biên độ dao động của vật \textit{The amplitude of the object's oscillation}
    \item Điều kiện kích thích ban đầu \textit{The initial stimulus conditions}
    \textbf{\item Khối lượng vật nặng \textit{The mass of the object}}
\end{enumerate}
\end{tcolorbox}

\begin{figure}
    \centering
    \includegraphics[width=1\linewidth]{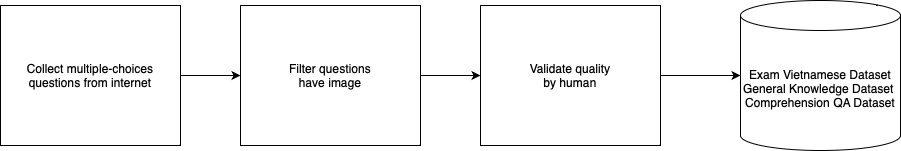}
    \caption{Process of collecting, rechecking, and creating multiple-choice data}
    \label{fig:multiple-choice-pipeline}
\end{figure}

\subsection{\textbf{General Knowledge Vietnamese Dataset}}

\subsubsection{Data Sources}

The General Knowledge dataset of ViLLM-Eval draws from a wide array of popular Vietnamese television quiz shows, such as "Who Wants to Be a Millionaire" and "Road to the Olympia Peak." The questions are collected from reputable online platforms known for their quality quiz and trivia content, guaranteeing the authenticity and diversity of the questions.

\subsubsection{Data Processing}

Once compiled, the questions are carefully formatted into a consistent multiple-choice structure, similar to MMLU \cite{hendrycks2021measuring}, each offering four potential answers. This conversion process is meticulously conducted to maintain the integrity and challenge of the original questions while ensuring they are understandable and relevant. Following the initial processing, a crucial step of human validation is undertaken, where subject matter experts review each question and answer for accuracy, clarity, and fairness. This human oversight ensures that the questions not only reflect a broad spectrum of general knowledge but are also free from errors and ambiguities, thereby upholding the high standards set for the ViLLM-Eval suite.

\begin{tcolorbox}[colback=white,colframe=black,boxrule=0.25pt]
\selectlanguage{vietnamese}
\textbf{Example:} Đến năm 2012, ai là người được phong đại tướng trẻ nhất Việt Nam?
\textit{By 2012, who was the youngest person to be promoted to the rank of general in Vietnam?}

\begin{enumerate}[label=(\alph*)]
    \item Nguyễn Chí Thanh
    \textbf{\item Võ Nguyên Giáp}
    \item Văn Tiến Dũng
    \item Chu Huy Mân
\end{enumerate}
\end{tcolorbox}

\subsection{\textbf{Comprehension QA Vietnamese Dataset}}

\subsubsection{Data Sources}

The Comprehension QA component is designed to assess deep reading comprehension and the ability to infer answers from complex texts. The data for this dataset is sourced from a variety of Vietnamese literary works, academic articles, and informative essays found on credible online platforms.. These texts are chosen for their depth of content, complexity of themes, and the richness of their linguistic expressions, providing a robust foundation for evaluating the comprehension skills of language models.

\subsubsection{Data Processing}

In this data set, these data are collected from competency assessment exams, and filtered for questions that require reading comprehension ability, each data includes a paragraph from 5 to 20 sentences long, one Multiple choice questions with 4 answers related to that passage. The format of this multiple-choice question is similar to that of MMLU \cite{hendrycks2021measuring}. Following the assembly of the questions, a human validation phase is integral to the process. Expert reviewers carefully assess each text and corresponding question set to ensure that the questions are meaningful, the correct answers are indisputably supported by the text, and the distractors are plausible yet clearly incorrect. This human validation process is essential for ensuring the reliability and validity of the Comprehension QA dataset, reinforcing ViLLM-Eval's commitment to providing a thorough and accurate assessment of language models' abilities.

\begin{tcolorbox}[colback=white,colframe=black,boxrule=0.25pt]
\selectlanguage{vietnamese}
\textbf{Example:} Trong một quần xã rừng tự nhiên ở vùng Đông Nam Á, các loài động vật ăn có cỡ lớn như bò rừng mỗi khi di chuyển thường đánh động và làm các loài côn trùng bay khỏi tổ. Lúc này, các loài chim như diệc bạc sẽ bắt các côn trùng bay khỏi tổ làm thức ăn. Việc côn trùng bay khỏi tổ cũng như việc chim diệc bạc bắt côn trùng không ảnh hưởng gì đến đời sống bò rừng. Chim gõ bò có thể bắt ve bét trên da bò rừng làm thức ăn. Mối quan hệ giữa diệc bạc và côn trùng dẫn đến hiện tượng gì? \textit{In a natural forest community in Southeast Asia, large herbivores like wild cattle often disturb and flush insects out of their nests while moving. During this time, birds such as the silver pheasant will catch the flying insects as food. The flushing of insects out of their nests as well as the hunting behavior of silver pheasants does not affect the life of wild cattle. Woodpeckers can catch beetles on the skin of wild cattle as food. What phenomenon does the relationship between silver pheasants and insects lead to?}

\begin{enumerate}[label=(\alph*)]
    \item Cạnh tranh khác loài \textit{Interspecies competition}
    \textbf{\item Khống chế sinh học \textit{Biological control}}
    \item Nhập cư- di cư \textit{Immigration-emigration}
    \item Hiệu ứng thắt cổ chai \textit{Bottleneck effect}
\end{enumerate}

\end{tcolorbox}

\begin{tcolorbox}[colback=white,colframe=black,boxrule=0.25pt]
\selectlanguage{vietnamese}
\textbf{Example:} Dựa vào thông tin sau để trả lời câu hỏi: Vào đầu thế kỉ XX, cùng với sự xuất hiện tầng lớp tư sản và tiểu tư sản, nhiều Tân thư, Tân báo của Trung Hoa cổ động cho tư tưởng dân chủ tư sản được đưa vào nước ta. Các sĩ phu yêu nước thức thời đã tiếp cận tư tưởng đó một cách nồng nhiệt. Những đổi mới của Nhật Bản sau cuộc Duy Tân Minh Trị (1868) càng củng cố niềm tin của họ vào con đường cách mạng tư sản. Đây chính là những điều kiện xã hội và tâm lí làm nảy sinh, thúc đẩy phong trào yêu nước theo khuynh hướng mới, trong đó, Phan Bội Châu, Phan Châu Trinh là những nhân vật tiêu biểu. Phan Bội Châu quê ở Nam Đàn, tỉnh Nghệ An. Ông là người chủ trương dùng bạo lực để giành độc lập. Năm 1902, ông lên đường vào Nam, rồi ra Bắc, tìm cách liên kết với những người cùng chí hướng. Tháng 5-1904, tại Quảng Nam, Phan Bội Châu cùng các đồng chí của ông thành lập Hội Duy Tân, chủ trương đánh đuổi Pháp, giành độc lập, thiết lập một chính thể quân chủ lập hiến ở Việt Nam. (SGK Lịch sử 11, trang 140-141) Sự kiện nào trên thế giới khiến các nhà yêu nước Việt Nam quyết định theo khuynh hướng tư sản?
\textit{Based on the following information, answer the question: At the beginning of the 20th century, along with the emergence of the bourgeoisie and petty bourgeoisie, many new books and newspapers from China promoting bourgeois democratic thought were brought into our country. Patriotic and progressive scholars embraced these ideas with enthusiasm. Japan's modernization after the Meiji Restoration (1868) further strengthened their belief in the path of bourgeois revolution. These were the social and psychological conditions that gave rise to and promoted the patriotic movement with a new tendency, in which Phan Boi Chau and Phan Chau Trinh were typical figures. Phan Boi Chau was born in Nam Dan, Nghe An province. He advocated using violence to achieve independence. In 1902, he went south, then north, seeking to connect with like-minded people. In May 1904, in Quang Nam, Phan Boi Chau and his comrades established the Duy Tan Hoi (Reformation Society), advocating the expulsion of the French, independence, and the establishment of a constitutional monarchy in Vietnam. (History textbook 11, pages 140-141) What event in the world made Vietnamese patriots decide to follow the bourgeois path?}

\begin{enumerate}[label=(\alph*)]
    \item Cách mạng Tân Hợi \textit{Tan Hoi Revolution}
    \textbf{\item Duy Tân Minh Trị \textit{Duy Tan Minh Tri}}
    \item Chiến tranh Nga-Nhật \textit{Russo-Japanese War}
    \item Cách mạng tháng Mười Nga \textit{October Revolution in Russia}
\end{enumerate}

\end{tcolorbox}

\subsection{Dama-2 7B}

In this work, we are thrilled to introduce Dama-2 7b \footnote{\href{https://huggingface.co/vietgpt/dama-2-7b}{https://huggingface.co/vietgpt/dama-2-7b}}, the Vietnamese language model built upon the renowned LLaMA architecture \cite{touvron2023llama}. Dama, short for Đà Mã (Vietnamese for LLaMa), this foundational model underwent a comprehensive training process from scratch, utilizing a substantial dataset of 270GB, which comprises English and Vietnamese texts extracted from the OSCAR corpus \cite{2022arXiv220106642A}. Notably, this extensive training set deliberately excludes any data from Wikipedia to ensure the most objective evaluation results on the LAMBADA dataset.

\begin{table}[h]
    \centering
    \begin{tabular}{|c|c|c|} \hline 
         &  LLaMa 2 7B& Dama-2 7B\\ \hline 
         tokenizer&  BPE& Byte-level BPE\\ \hline 
         $d_{model}$&  4096 & 4096\\ \hline 
         $n_{layers}$&  32 & 32\\ \hline 
 $n_{heads}$& 32 & 32\\ \hline 
 vocab\_size& 32,000 & 50,261 \\ \hline 
 context\_length& 4,096 & 4,096 \\ \hline
    \end{tabular}
    \caption{This table compares the important configurations of two large language models, LLaMa-2 7B and Dama-2 7B, including the tokenizer type, model dimension, number of layers, number of attention heads, vocabulary size, and context length for each model. This comparison provides insights into the architecture and scale of the two models.}
    \label{tab:model_config_comparison}
\end{table}

In the subsequent phase of development, the model was subject to a meticulous fine-tuning process, employing a variety of publicly available Supervised Fine-Tuning (SFT) datasets. This stage was instrumental in adapting the core capabilities of the LLaMA-based model to cater to more specific tasks and domains, thereby significantly elevating its performance and versatility in addressing a broad spectrum of challenges within the field of natural language processing.

\section{Evaluation}

This work aims to use perplexity and accuracy as the metric. While ground-truth labels of the validation splits are released, the labels of the test split are kept private. This is to ensure the fair use of the ViLLM-Eval, as the ViLLM-Eval data may unconsciously be included in pre-training data due to web crawling. Instead, users can utilize our provided library and run the evaluation directly, where the ground-truth labels are embedded within the library but encrypted. A public leaderboard is maintained to display the test accuracy. Users have the option to include their submission results in the live leaderboard, depending on their own preference.

For the LAMBADA Vietnamese dataset. The evaluation of LLMs involves calculating the perplexity of the last word of the last sentence of each input paragraph. Additionally, accuracy is used to show that the model accurately predicts that last word. At the same time, accuracy also helps calculate average comparison results.

For datasets in the form of multiple-choice questions, LLMs are evaluated in a five-shot setting on ViLLM-Eval, where the five exemplars are from the development split. Probability distributions are utilized to determine the LLM's predictions. If one of the four answer choices has the highest probability, it is considered the LLM's predicted answer \cite{brown2020language}. A zero-shot setting yields suboptimal results for foundational models that have only been pretrained, as they lack specific fine-tuning for the task. Therefore, the five-shot setting is employed. This few-shot approach aims to guide the model and provide it with the necessary context to better understand the task and the expected format for the responses.

However, some models' inference APIs do not provide probability distributions. In such cases, the predicted answer is extracted based on the model's generated output. Regular expressions are employed as the extraction method to reliably obtain the models' answer choices from their generated responses. For the five-shot setting, where the models are provided with a few examples to guide their understanding, only the answer-only results are presented. This focus on answer-only results is due to the tendency of the model's predictions to follow specific patterns facilitated by the few-shot examples, enabling reliable extraction of the answer choices.

\newpage
\section{Experiment}

\subsection{Models}

This work aims to present a comprehensive overview of accessible, top-tier LLMs that support Vietnamese input in the context of evaluating Large Language Models (LLMs) for Vietnamese language processing. This evaluation includes a diverse range of models from various organizations, differing in scale. From OpenAI, ChatGPT GPT-3.5-turbo \cite{chatgpt} is included, representing one of the advanced GPT model variants tailored for efficient and responsive interactions in multiple languages, including Vietnamese.

ViLLM-Eval focuses on LLMs designed specifically for Vietnamese, reflecting the efforts of local and regional developers to advance language technology. Recent LLMs developed by Vietnamese organizations or individuals such as Vistral-7B-Chat \cite{chien2023vistral}, PhoGPT-4B-Chat \cite{PhoGPT}, VinaLLaMA-7B-Chat \cite{nguyen2023vinallama}, SeaLLM-7B-v2 \cite{damonlpsg2023seallm}, and other models evaluated during the VLSP 2023 VLLM conference are also included.

Vistral-7B-Chat \cite{chien2023vistral} is an extension of the Mistral 7B model, continually pre-trained on a diverse and carefully curated dataset of Vietnamese texts, and further fine-tuned on instructional data to align the model with safety criteria in Vietnam.

PhoGPT-4B-Chat \cite{PhoGPT} is a state-of-the-art 3.7 billion parameter model, employing the latest advances in large language model architectures and training techniques. It was pre-trained from scratch on a massive Vietnamese corpus of 102 billion tokens.

SeaLLM-7B-v2 \cite{damonlpsg2023seallm} is the state-of-the-art 7 billion parameter multilingual large language model tailored for Southeast Asian languages, including Vietnamese. It offers superior performance at half the model size across a diverse array of multilingual tasks, such as world knowledge, mathematical reasoning, and instruction following.

We hope that developers of other potent Vietnamese LLMs will engage with our platform, submitting and, if they choose, publicizing their models' results. This collaborative effort would not only enrich our understanding but also foster advancements in the development of LLMs tailored for the Vietnamese language and its diverse dialects.

\begin{table}[h]
    \centering
    \begin{tabular}{|c|c|c|c|c|} \hline 
         Model&  Creator&  VLSP participatent&  \#Parameters& Access\\ \hline 
         ChatGPT&  OpenAI&  No&  Unknown& API\\ \hline 
         Vistral-7B-Chat&  Viet-Mistral&  No&  7B& Weight\\ \hline 
         VinaLLaMA-7B-Chat&  VILM&  No&  7B& Weight\\ \hline 
         PhoGPT-4B-Chat&  VinAI&  No&  4B& Weight\\ \hline 
         SeaLLM-7B-v2&  SeaLLMs&  No&  7B& Weight\\ \hline
 Dama-2-7B& VietGPT& No& 7B&Weight\\\hline
 greennode-14b& GreenNode.ai& Yes& 14B&API\\\hline
 greennode-7b& GreenNode.ai& Yes& 7B&API\\\hline
 cpt-smartbot-13b& CPT\_Smartbot& Yes& 13B&API\\\hline
 ura-llama-13b& ura-hcmut& Yes& 13B&API\\\hline
    \end{tabular}
    \caption{Overview of language models evaluated in this study, including their creators, participation in the VLSP competition, number of parameters, and the mode of access to the models. This table provides a concise comparison of the models' backgrounds and accessibility for further research or application development.}
    \label{tab:my_label}
\end{table}
\subsection{Results}

The LAMBADA dataset challenge was used to assess the models' ability in text understanding and prediction, where a lower perplexity (PPL) indicates better performance. The accuracy on LAMBADA (LAMBADA acc) measures how often the models correctly predict the final word in a passage. For the Exam, General Knowledge, and Comprehension QA tasks, the models are evaluated based on their accuracy in answering multiple-choice questions, with higher scores indicating better performance.

\begin{table}
\centering
\begin{tabular}{|l|c|c|c|c|c|} \hline
\textbf{Model} & \textbf{\makecell{LAMBADA \\ PPL}}& \textbf{\makecell{LAMBADA \\ acc}} & \textbf{Exam} & \textbf{Gen. Knowl.} & \textbf{Comp. QA}\\ \hline
ChatGPT & TBA & TBA & \underline{0.4862} & 0.5405 & \textbf{0.7421}\\ \hline
Vistral-7B & \textbf{8.4821} & \textbf{0.4482} & \textbf{0.4876} & \textbf{0.6600} & \underline{0.6978}\\ \hline
VinaLLaMA-7B & 29.2666 & 0.3021 & 0.3434 & 0.4360 & 0.6300\\ \hline
PhoGPT-4B & 12.2344 & 0.3802 & 0.4120 & 0.5215 & 0.6177\\ \hline
SeaLLM-7B-v2& 212.4439 & 0.2544 & 0.3439 & 0.4385 & 0.6333\\ \hline
 Dama-2-7B& \underline{11.7326} & \underline{0.397}& \underline{0.4266}& 0.5475&0.6722\\\hline
 greennode-14b& 29.5967& 0.3157& 0.3672& 0.4680&0.6711\\\hline
 greennode-7b& 31.6258
& 0.1845& 0.2892& 0.3335&0.6122\\\hline
 cpt-smartbot-13b& 21.9864& 0.2992& 0.3473& 0.4455&0.6633\\\hline
 ura-llama-13b& 17.5614& 0.3263& 0.3420& 0.438&0.6556\\\hline
\end{tabular}
\caption{Comparative performance of language models on NLP tasks. This includes the LAMBADA dataset (PPL and accuracy), Exam, General Knowledge (Gen. Knowl.), and Comprehension QA (Comp. QA). ChatGPT's LAMBADA metrics are unavailable due to API limitations.}
\label{tab:performance_comparison}
\end{table}
Upon examining the results, it is evident that ChatGPT outperforms its competitors across all tasks for which scores are reported. It is noteworthy that the LAMBADA metrics for ChatGPT are missing, which is attributed to the model's operational mechanism via API. Unlike traditional models that provide probability distributions essential for evaluating perplexity, ChatGPT's inference APIs do not furnish such data, rendering the evaluation of LAMBADA metrics infeasible.

Vistral-7B stands out with commendable performance across all tasks, especially achieving the lowest perplexity on the LAMBADA dataset and notable accuracy on the General Knowledge task. Dama-2-7B and PhoGPT-4B present competitive performances, with Dama-2-7B slightly outperforming PhoGPT-4B in most tasks, except for Comprehension QA. Despite SeaLLM-7B-v2 potentially grappling with higher perplexity, it's worth noting that its multilingual nature might contribute to this aspect. Nevertheless, it still demonstrates competitive accuracy on the Comprehension QA task.

These findings illuminate the distinct strengths and potential enhancement areas for each model, offering valuable insights into the present landscape of language model capabilities and guiding future research and development trajectories in the domain.

\section{Discussion}

This work aims to emphasize that assessing large language models (LLMs) should extend beyond the realm of simple chatbots and aid in equipping LLMs to operate in more intricate environments. This belief was the driving force behind the development of ViLLM-Eval, an evaluation suite designed to be challenging. It is optimistic that ViLLM-Eval has contributed significantly to advancements in this area, especially within the Vietnamese context. However, it's important to recognize that ViLLM-Eval, like all English-based benchmarks, is not flawless for evaluating LLMs. There exist numerous other capabilities, such as the ability to reason and interact with APIs, along with several dimensions that surpass mere accuracy, encompassing safety, bias, and resilience.

Moreover, the ever-evolving nature of language models and the dynamic linguistic landscape necessitate continuous refinement and adaptation of evaluation methodologies. As LLMs continue to evolve and incorporate new features and capabilities, evaluation frameworks must keep pace to ensure relevance and effectiveness.

In conclusion, the VLSP 2023 - Vietnamese Large Language Models shared task was formulated to propel research development in the domain of Large Language Models in Vietnamese. By addressing the limitations of current evaluation approaches and embracing a more holistic perspective, this work aims to unlock the full potential of LLMs and pave the way for their responsible and effective deployment in real-world applications.

\newpage
\section{Future Work}

The development and refinement of ViLLM-Eval mark an important step towards comprehensive evaluation of Vietnamese Large Language Models (LLMs). However, there are several avenues for future work and improvement:

\begin{itemize}
    \item \textbf{Expansion of Evaluation Tasks}: While ViLLM-Eval covers a diverse range of tasks, including comprehension, general knowledge, and exam-style questions, there is scope for further expansion. Future iterations could include tasks related to language generation, dialogue understanding, and real-world application scenarios to provide a more holistic evaluation of LLM capabilities.
    
    \item \textbf{Addressing Bias and Fairness}: As LLMs increasingly play a role in decision-making processes, it becomes crucial to evaluate their performance with respect to bias and fairness. Future work could focus on developing evaluation metrics and tasks specifically designed to assess LLMs' adherence to ethical principles and equitable treatment of diverse demographic groups.
    
    \item \textbf{Community Engagement}: Collaboration and engagement with the research community, industry stakeholders, and end-users are essential for the continued development and refinement of ViLLM-Eval. Encouraging participation in benchmarking efforts, soliciting feedback, and fostering a culture of transparency and openness will contribute to the ongoing improvement of evaluation practices.
    
    \item \textbf{Longitudinal Studies}: Tracking the performance of LLMs over time through longitudinal studies can provide valuable insights into their evolution and adaptation. Continuously updating ViLLM-Eval with new data and benchmarks will ensure its relevance and effectiveness in evaluating the latest advancements in Vietnamese LLM technology.
\end{itemize}

Overall, future work in these areas will contribute to the ongoing advancement of evaluation methodologies for Vietnamese LLMs, ultimately leading to more robust, reliable, and inclusive language technologies tailored to the needs of Vietnamese users.

\newpage
\section{Acknowledgements}
This work is supported by Intelligent Integration Co., Ltd (INT$^2$), Ha Noi, Vietnam and HPC SYSTEMS Inc., Tokyo, Japan. We would like to express to the organizers of the Vietnamese Language and Speech Processing (VLSP) for their exceptional dedication in orchestrating the Vietnamese Large Language Models challenge. Their efforts have been instrumental in fostering advancements in the field. We also would like to express our sincere gratitude to the students: Nguyen Trong Chi, Nguyen Phuoc Nguyen, Vo Huu Tri and Nguyen Minh Trung Hieu from the NLP-KD Lab, Ton Duc Thang University. They have contributed to the completion of this research. 

\newpage
\printbibliography 

\end{document}